\documentclass[letterpaper]{article} 
\usepackage{aaai2026}  
\usepackage{times}  
\usepackage{helvet}  
\usepackage{courier}  
\usepackage[hyphens]{url}  
\usepackage{graphicx} 
\usepackage{natbib}  
\usepackage{caption} 
\usepackage{amsmath}
\usepackage{amssymb}        
\usepackage[utf8]{inputenc} 
\usepackage[T1]{fontenc}    
\usepackage{xcolor}

\usepackage{paralist}
\setdefaultleftmargin{1em}{}{}{}{}{}

\definecolor{bluelightgray}{rgb}{0.4, 0.7, 0.8}
\definecolor{electriclime}{rgb}{0.8, 1.0, 0.0}
\definecolor{malachite}{rgb}{0.04, 0.85, 0.32}
\definecolor{darkred}{rgb}{0.55, 0.0, 0.0}
\definecolor{black}{rgb}{0.0, 0.0, 0}
\definecolor{darkblue}{rgb}{0.0, 0.0, 0.50}
\definecolor{darkgreen}{rgb}{0.0, 0.2, 0.13}
\definecolor{darkorchid}{rgb}{0.6, 0.2, 0.8}
\definecolor{neonskyblue}{rgb}{0.4,1,1}
\definecolor{neongreen}{rgb}{0.6,0,0.6}

\usepackage{changepage} 
\usepackage{algorithm}
\usepackage{algorithmic}
\usepackage[utf8]{inputenc}
\usepackage{enumitem}
\usepackage{caption}
\usepackage{subfigure}
\usepackage{multirow}
\usepackage{pbox}
\usepackage{booktabs}
\usepackage{graphicx}
\usepackage{bm}
\usepackage{siunitx}
\usepackage{lipsum}
\usepackage{mathtools}

\usepackage{tabularx}
\usepackage{arydshln}
\usepackage{nicefrac}       
\usepackage{microtype}      
\usepackage{xcolor}         
\usepackage{paralist}
\usepackage[table]{xcolor}

\usepackage{newfloat}
\usepackage{listings}
\DeclareCaptionStyle{ruled}{labelfont=normalfont,labelsep=colon,strut=off} 
\floatstyle{ruled}
\newfloat{listing}{tb}{lst}{}
\floatname{listing}{Listing}
\title{How Many Experts Are Enough? \\Towards Optimal Semantic Specialization for Mixture-of-Experts}
\author{
    Sumin Park\textsuperscript{\rm 1}, 
    Noseong Park\textsuperscript{\rm 1}
}
\affiliations{
    \textsuperscript{\rm 1}Korea Advanced Institute of Science and Technology (KAIST)\\
    \{psmiz, noseong\}@kaist.ac.kr
}

\begin{document}

\maketitle

\begin{abstract}
Finding the optimal configuration of Sparse Mixture-of-Experts (SMoE) that maximizes semantic differentiation among experts is essential for exploiting the full potential of MoE architectures. However, existing SMoE frameworks either heavily rely on hyperparameter tuning or overlook the importance of diversifying semantic roles across experts when adapting the expert pool size.
We propose \textbf{M}ixture-of-Experts for \textbf{A}daptive \textbf{S}emantic \textbf{S}pecialization (\textbf{\textsc{MASS}}), a semantic-aware MoE framework for adaptive expert expansion and dynamic routing. \textsc{MASS} introduces two key advancements: (i) a gradient-based semantic drift detector that prompts targeted expert expansion when the existing expert pool lacks capacity to capture the full semantic diversity of the data, and (ii) an integration of adaptive routing strategy that dynamically adjusts expert usage based on token-level routing confidence mass.
We first demonstrate that \textsc{MASS} reliably converges to the point of optimal balance between cost-performance trade-off with notably improved sematic specialization in a highly controlled synthetic setup. Further empirical results on real-world datasets across language and vision domains show that \textsc{MASS} consistently outperforms a range of strong MoE baselines, demonstrating its domain robustness and enhanced expert specialization.
\end{abstract}

\section*{Introduction}

The Sparse Mixture-of-Experts (SMoE) approach has recently emerged as an effective mechanism that substantially improves model capacity with minimal computational overhead~\citep{shazeer2017outrageouslylargeneuralnetworks,lepikhin2020gshardscalinggiantmodels,fedus2022switchtransformersscalingtrillion}. Incorporating SMoE structure into deep neural networks, particularly large-scale Transformer-based architectures, has significantly boosted performance across diverse tasks and modalities~\citep{li2023sparsemixtureofexpertsdomaingeneralizable, qiu2024emergent}. A fundamental advantage of SMoE is its inherent scalability and flexibility, achieved by selectively routing input tokens to a sparse subset of specialized subnetworks (experts). This sparsity-based selective routing mechanism is especially appealing for large-scale neural architectures, such as Large Language Models (LLMs), which commonly struggle with substantial computational demands, handling massive model sizes and diverse datasets~\cite{zhu2024llamamoebuildingmixtureofexpertsllama, yang2024qwen2technicalreport, jiang2024mixtralexperts,dai2024deepseekmoeultimateexpertspecialization}.

However, a crucial issue when employing SMoE lies in determining the optimal size of the expert pool, commonly denoted as \( K \), and the number of active experts per token, referred to as top-\( k \) selection~\citep{shazeer2017outrageouslylargeneuralnetworks, fedus2022switchtransformersscalingtrillion, zhou2022mixtureofexpertsexpertchoicerouting, zhou2024brainformerstradingsimplicityefficiency}. Current methods remain largely heuristic and reliant on resource-intensive hyperparameter searches. 

Recently, \textbf{DynMoE}~\cite{guo2025dynamicmixtureexpertsautotuning} has been proposed as a promising solution that adaptively adjusts the expert set size based on token-expert routing coverage. However, DynMoE relies solely on token-level expert activation pattern and does not explicitly assess whether the current expert pool becomes functionally saturated to capture semantic variations in input data.
This overlooks a critical aspect of MoE modeling: fine-grained semantic specialization across experts, which is essential for minimizing functional redundancy in experts and thus, promoting complementary mixture of subnetworks.

\begin{figure*}[h]
    \centering
    \includegraphics[width=0.90\textwidth]{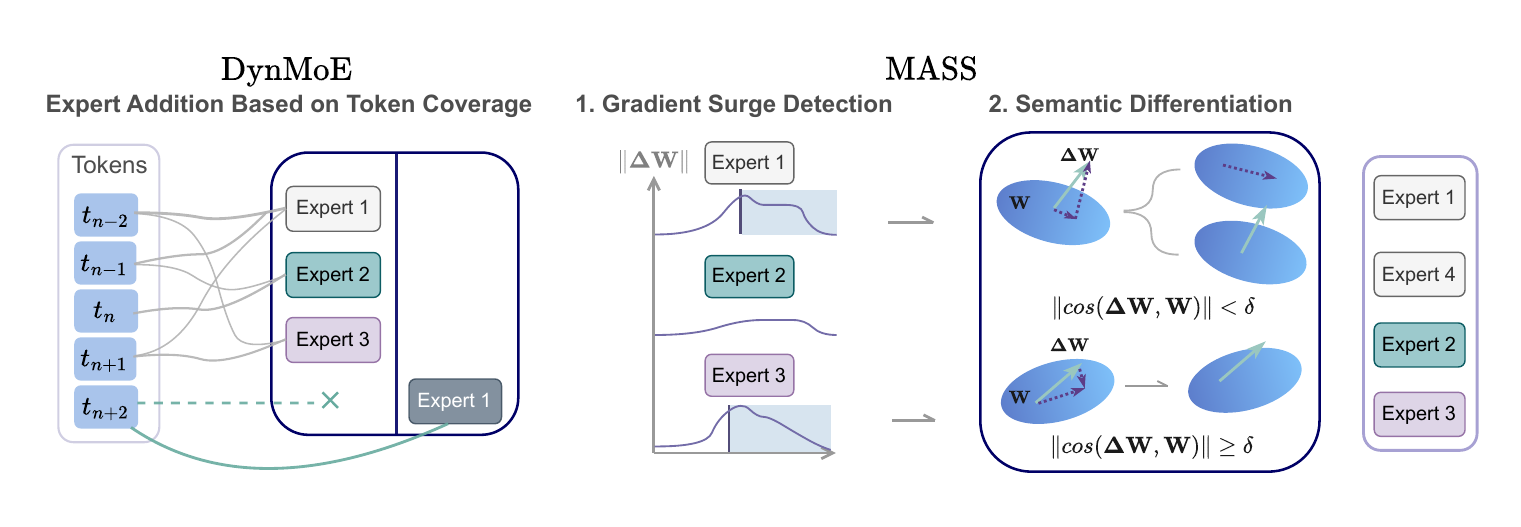}
    \vspace{-0.5em}
    \caption{
    {Comparison of Adaptive Expert Expansion Mechanisms.}
    {Left (DynMoE):} Experts are added when tokens fail to activate any existing expert, indicating insufficient token coverage. This strategy relies on token-level activation statistics, without explicitly assessing semantic adequacy.
    {Right (MASS):} Expert expansion is triggered through a two-step process: (1) MASS statistically detects upward shifts in gradient norm over training steps to identify experts under high update pressure, and (2) it evaluates semantic drift by measuring misalignment between the gradient update \( \Delta \mathbf{W} \) and the expert’s current representation \( \mathbf{W} \). If the alignment \( \|\cos(\Delta \mathbf{W}, \mathbf{W})\| \) falls below a threshold \( \delta \), a new expert is instantiated to specialize on the diverging semantics.
    }
    \label{fig:dynmoevsmass}
\end{figure*}


To address this limitation, we propose \textbf{\textsc{MASS}} (\textbf{M}ixture-of-Experts for \textbf{A}daptive \textbf{S}emantic \textbf{S}pecialization), a semantic-aware MoE framework that adaptively expands the expert pool based on gradient-based indicators of functional insufficiency. MASS identifies when an expert is overloaded beyond its semantic capacity by combining two signals: (1) a sustained increase in gradient magnitude detected by probabilistic \textit{Change Point Detection (CPD)} approach, and (2) a semantic drift indicated by misalignment between the current gradient update and the expert’s historical weight representation. If both conditions are met, a new expert is instantiated to specialize on the emerging role.
{Figure~\ref{fig:dynmoevsmass}} provides a visual comparison of expert expansion mechanism between DynMoE and MASS. While DynMoE heuristically expands experts based on unassigned tokens, MASS uses a principled two-stage mechanism that directly targets semantic insufficiency, enabling finer-grained and functionally grounded expert specialization.

We complement this expansion mechanism with an integration of a probabilistic 
routing strategy inspired by Top-$p$ routing~\cite{topprouting}. Instead of 
selecting a fixed number of experts (as in top-$k$ gating), the gating module 
activates the minimal subset of experts whose cumulative routing probability 
exceeds a confidence threshold $p$. Under this strategy, confident tokens naturally use fewer experts, reducing computation, while uncertain tokens are routed
to a larger set of experts for richer processing. This yields a balanced and adaptive 
expert allocation that complements MASS’s principled expansion criterion.

To validate the effectiveness of our approach, we first rigorously tested {\textsc{MASS}} in a controlled synthetic setting with language-like data modeled by multinomial Hidden Markov Models (HMMs), following the same settings in GINC dataset generation ~\citep{xie2022explanationincontextlearningimplicit}. This setup allows us to examine whether \textsc{MASS} can adaptively converge to an optimal balance on the trade-off between the performance and computational cost, i.e., the number of experts. Beyond synthetic validation, we further evaluate {\textsc{MASS}} on real-world datasets spanning both language and vision tasks in comparison to strong MoE baselines, including DynMoE. 

\paragraph{Contributions} Our main contributions are as follows:
\begin{itemize}
    \item We propose {\textsc{MASS}}, a semantic-aware adaptive MoE expansion framework that expands the expert pool using gradient-based semantic drift detection.
    \item We adopt a confidence-driven Top-$p$ routing, enabling sufficient and balanced per-token expert attention.
    \item We empirically validate \textsc{MASS} on both synthetic and real-world tasks, demonstrating its ability to determine the optimal expert pool size with consistent performance gains over diverse baselines across multiple domains.
\end{itemize}

\section*{Background}

\paragraph{Sparse Mixture of Experts}  

Recent works have actively employed Sparse Mixture-of-Experts (SMoE) structure in modern deep learning architectures, demonstrating its wide applicability across multiple domains~\citep{li2023sparsemixtureofexpertsdomaingeneralizable, qiu2024emergent, puigcerver2024sparsesoftmixturesexperts}. 
SMoE layers are commonly integrated into Transformer-based architectures by replacing the standard feed-forward network with a sparse, gated mixture of expert sub-networks. A gating module routes each token to a small subset of experts, typically via top-$k$ token-wise selection, enabling increased model capacity with reduced computational cost~\citep{fedus2022switchtransformersscalingtrillion, dai2024deepseekmoeultimateexpertspecialization}.

Beyond foundational design, SMoE has also been explored in various fine-tuning scenarios~\cite{zadouri2023pushingmixtureexpertslimit, li2024mixloraenhancinglargelanguage,
wu2024omnismolaboostinggeneralistmultimodal}. Among them, MoEfication~\citep{zhang2022moeficationtransformerfeedforwardlayers} restructures pre-trained feed-forward layers into sparse expert modules for efficient inference and GMoE~\citep{li2023sparsemixtureofexpertsdomaingeneralizable}, initialized from a pretrained vision transformers, enhance domain generalization in vision tasks.

\paragraph{Multinomial HMM}  
The Hidden Markov Model (HMM)~\citep{baum1966statistical} is a classical generative model designed for sequential data, where a sequence of observable outputs is generated by an underlying sequence of latent variables that evolve according to a first-order Markov process. In this framework, each latent state generates an observation independently based on a state-dependent emission distribution. Let $\{h_t\}_{t=1}^T$ represent the sequence of latent states and $\{o_t\}_{t=1}^T$ the corresponding observed outputs. The model is governed by three components: an initial state distribution $\pi = p(h_1)$, a state transition function $\mathcal{T} = p(h_t \mid h_{t-1})$, and an emission distribution $\mathcal{E} = p(o_t \mid h_t)$. The joint probability of the full sequence is expressed as:
\[
p(o_{1:T}, h_{1:T} \mid \theta) = \pi(h_1) \prod_{t=2}^T \mathcal{T}(h_t \mid h_{t-1}) \prod_{t=1}^T \mathcal{E}(o_t \mid h_t),
\]
where $\theta = \{\pi, \mathcal{T}, \mathcal{E}\}$ denotes the full set of model parameters. A {categorical HMM} is a variant where each hidden state emits observations drawn from a categorical distribution over a discrete set of outcomes. Each latent state $h_t \in \{1, \dots, N\}$ is tied to a categorical distribution over a finite observation set $\mathcal{O} = \{1, \dots, M\}$. The emission probabilities are parameterized by a matrix $\mathcal{E} \in \mathbb{R}^{N \times M}$, where each row represents the probability of producing each outcome at a given hidden state. Along with a transition matrix $\mathcal{H} \in \mathbb{R}^{N \times N}$ and an initial state distribution $\pi \in \mathbb{R}^N$, this model effectively captures sequential patterns in symbolic data streams.

\paragraph{Probabilistic Change Point Detection}  
Change Point Detection (CPD) refers to the task of identifying abrupt distributional shifts in time series. A classical approach relies on the Cumulative Sum (CUSUM) algorithm, 
which tracks cumulative sum of deviations from a reference distribution and triggers a change point when this drift exceeds a fixed threshold. Probabilistic CPD builds on this by recasting the drift detection as a statistical test. Instead of fixed thresholds, it quantifies how likely the observed cumulative deviation is under the null hypothesis of no distributional change.

Formally, given a time series \( \{X_t\}_{t=1}^T \), we standardize observations as 
$Z_t = \frac{X_t - \hat{\mu}_X}{\hat{\sigma}_X}$,
where \( \hat{\mu}_X \) and \( \hat{\sigma}_X \) are estimates of the mean and standard deviation of the series. The cumulative sum 
$S_T = \sum_{t=1}^T Z_t$
approximates \( \mathcal{N}(0, T) \) under the assumption that the $Z_t$ is i.i.d. and satisfies the Central Limit Theorem (CLT) conditions
, yielding a normalized statistic 
\[
\tilde{S}_T = {S_T}/\sqrt{T} \sim \mathcal{N}(0,1).
\]
Given a test statistic \(\tilde{s}_T\) of the random variable \(\tilde{S}_T\), 
the probability of observing a value less than or equal to \(\tilde{s}_T\) is as:
\[
\Phi(\tilde{s}_T) \approx P(\tilde{S}_T \leq \tilde{s}_T),
\]
where \( \Phi(\cdot) \) denotes the cumulative distribution function (c.d.f.) of the standard normal distribution.
The right-tail probability \( 1 - \Phi(\tilde{s}_T) \) serves as a p-value in hypothesis testing. A change point is flagged when it falls below a predefined threshold, enabling robust and underlying distribution-free detection of drift.

\section{Method}\label{sec:method}

\subsection{Overview of \textsc{MASS}}
In this section, we provide full algorithmic details of \textsc{MASS}.
\begin{enumerate}
    \item \textbf{Adaptive Expert Expansion.} During the early phase of training, \textsc{MASS} dynamically adjusts the expert pool size by detecting semantic drift through gradient-based monitoring. 
    \item \textbf{Standard Training.} The second phase fixes the expert set derived from the first phase and performs conventional supervised training.
\end{enumerate}
\noindent Throughout the entire training process, expert routing is governed by a {routing-mass-based adaptive {Top-$p$}} strategy. 

\subsection{MoE Architecture}
We define the structure of the MoE layer shared across all subsequent experiments. The layer consists of a token-wise gating module and a pool of \( K \) experts, where \( K \) denotes the current expert pool size at a given point in training. For each input token representation \( \mathbf{x} \in \mathbb{R}^d \), a subset of experts \( \mathcal{N}(\mathbf{x}) \subseteq \{1, \dots, K\} \) is selected by the router, and the output is computed as:
\[
\mathbf{y} = \sum_{k \in \mathcal{N}(\mathbf{x})} r_k(\mathbf{x}) \cdot e_k(\mathbf{x}).
\]
Here, \( e_k(\cdot) \) denotes the \( k \)-th expert function, and \( r_k(\mathbf{x}) = [\mathbf{r}(\mathbf{x})]_k \) is the routing score assigned to expert \( k \). The full routing distribution is computed as \( \mathbf{r}(\mathbf{x}) = \text{Softmax}(\mathbf{x}^\top \mathbf{W}_r) \), where \( \mathbf{W}_r \in \mathbb{R}^{d \times K} \) is the gating weight matrix.

\paragraph{Routing-Mass Based Expert Selection}
To determine $\mathcal{N}(\mathbf{x})$ per token, the model adopts the {Top-$p$} strategy. Let $r^{(1)} \geq r^{(2)} \geq \cdots \geq r^{(K)}$ denote the sorted routing scores in descending order, and let $\mathcal{I}_j$ be the original expert index corresponding to the $j$-th largest score $r^{(j)} = r_{\mathcal{I}_j}(\mathbf{x})$. The number of active experts $k^\ast$ is chosen as the smallest index such that:
\[
\sum_{j=1}^{k^\ast} r^{(j)} \geq p,\quad \mathcal{N}(\mathbf{x}) = \{\mathcal{I}_1, \dots, \mathcal{I}_{k^\ast}\},
\]
where $p \in (0, 1)$ is a hyperparameter controlling the routing confidence threshold. Instead of assigning a hard number of experts to every token regardless of certainty, this confidence-aware strategy dynamically allocates expert capacity according to the model’s routing uncertainty.

\subsection{Adaptive Expert Expansion}
Training begins with an initial number of experts $K_{\text{init}}$ and expands the expert pool up to a maximum expert limit $K_{\max}$ during the first 10\% of total training steps. 

\paragraph{Gradient Monitoring via Probabilistic CPD}
For each expert $e_k$ parameterized by $\theta_k$, the L2 norm of the task gradient is tracked at each training step $t$ as 
\[
g^{(k)}_t := \|\nabla_{\theta_k} \mathcal{L}_t\|_2,
\]
where $\mathcal{L}_t$ denotes the loss at step $t$. To ensure stability, each expert undergoes a warmup phase of $T_\text{warmup}$ steps before any CPD statistics are computed. After warmup, a sliding window of size $\omega$ maintains the $\omega$-most recent gradient norms:
\[
\mathcal{G}^{(k)}_t := \{ g^{(k)}_{t-\omega+1}, \dots, g^{(k)}_t \}.
\]
Within this window, the sample mean and unbiased standard deviation are computed as
\[
\mu_t^{(k)} := \frac{1}{\omega} \sum_{g \in \mathcal{G}_t^{(k)}} g, \quad
\sigma_t^{(k)} := \sqrt{{\frac{1}{\omega - 1} \sum_{g \in \mathcal{G}_t^{(k)}} \left( g - \mu_t^{(k)} \right)^2 }}.
\]
The cumulative z-score within window is then computed as:
\[
s^{(k)}_t := \sum_{i=t-\omega+1}^t z^{(k)}_i, \quad
z^{(k)}_t := \frac{g^{(k)}_t - \mu^{(k)}_t}{\sigma^{(k)}_t}.
\]
yielding the normalized test statistic: $\tilde{s}^{(k)}_t := {s^{(k)}_t}/{\sqrt{\omega}}$.

\noindent Assuming normality under the null hypothesis, i.e., $\tilde{s}^{(k)} \sim \mathcal{N}(0,1)$, the corresponding right-tailed $p$-value is given by:
\[
p^{(k)}_t := 1 - \Phi(\tilde{s}^{(k)}_t),
\]
where $\Phi(\cdot)$ denotes the cumulative distribution function of the standard normal distribution. To decide whether to reject the null hypothesis of no distributional shift, we compare the $p$-value against a predefined significance level $\alpha \in (0, 1)$, and reject when $p^{(k)}_t \leq \alpha$. This right-tailed test identifies statistically significant upward shifts in the gradient magnitude distribution, indicating that expert $e_k$ may require substantial adaptation in its representation.

\paragraph{Semantic Alignment Test}
For a flagged expert $e_k$ by prior gradient monitoring, semantic misalignment is diagnosed by evaluating the cosine similarity between the current gradient matrix $\nabla^{(k)}$ and the expert’s weight matrix $\mathbf{W_e}^{(k)}$:
\[
\cos(\nabla^{(k)}, \mathbf{W_e}^{(k)}) := \frac{\langle \nabla^{(k)}, \mathbf{W_e}^{(k)} \rangle}{\|\nabla^{(k)}\|_2 \cdot \|\mathbf{W_e}^{(k)}\|_2},
\]
where both matrices are flattened into vectors before computing similarity. Here, \( \mathbf{W_e}^{(k)} \) refers to the projection weight of the first linear layer in the expert function. If the similarity satisfies \( \|\cos(\nabla^{(k)}, \mathbf{W_e}^{(k)})\| < \delta \), with \( \delta = 0.001 \) as a fixed threshold indicating effective orthogonality, the expert is considered to have undergone semantic drift, prompting the addition of a new expert for diverging semantic role.

\paragraph{Expert Duplication with Gradient Decomposition}
Upon confirming a semantic drift by two-step detection, the expert $e_k$ is duplicated to form a new expert, denoted as $e_k'$. $e_k'$ inherits the full gradient update, while the original expert $e_k$ receives only the component aligned with its current weight representation. 

An analogous update is applied to the gating network. Let \( \mathbf{w}_k = [\mathbf{W}_r]_{:,k} \) denote the gating column vector corresponding to expert \( e_k \). To support the new expert \( e_k' \), a corresponding gating vector \( \mathbf{w}_{k}' \) is initialized by cloning:
$\mathbf{w}_{k}' \leftarrow \mathbf{w}_{k}$.
Then, the original gating vector receives only the aligned gradient, while the new one takes the full gradient. 

This selective update helps preserve the original semantics of $e_k$ while allowing $e_k'$ to explore a divergent role. To mitigate redundancy afterwards, a regularization term is applied to encourage divergence in their routing behavior.

\paragraph{Redundancy Regularization}
Following expert duplication, experts and their gating vectors are initialized to be nearly identical, which risks the functional collapse of the duplicated experts, presenting redundant behavior.
To mitigate this, we introduce a regularization term that penalizes high alignment between the gating vectors of duplicated expert pairs. 
Formally, the regularization loss is defined as:
\[
\mathcal{L}_{\text{red}} =  \frac{1}{|\mathcal{P}|} \sum_{(i,j) \in \mathcal{P}} \left( \cos(\mathbf{w}_i, \mathbf{w}_j) \right)^2,
\]
where $\mathcal{P}$ is the set of all duplicated expert pairs, and $\mathbf{w}_i$ denotes the gating vector of expert $i$. By penalizing similarity in gating vectors, we encourage input tokens to be routed differently between the pair, promoting eventual divergence in their learned semantic roles. 
Note that this regularization is only applied during the expansion phase to avoid severe distortion of natural evolution of the routing behavior.

\paragraph{Expansion Stopping Criteria}

\begin{algorithm}[tb]
\caption{NLL Comparison for Expansion Stopping}
\label{alg:nllstopping}
\begin{algorithmic}[1]
\REQUIRE Patience counter $\gamma > 0$, expert set $\mathcal{K}$, list of duplicated experts $\mathcal{P}$

\IF{the $(m{+}1)$-th expert has just been added \AND $|\mathcal{P}| \geq 2$}
    \STATE Let $e_{m}$ be the expert added in the $m$-th expansion step \
    \STATE Temporarily disable $e_{m}$ and compute the masked loss $\mathcal{L}_{\text{mask}}$
    \STATE Restore $e_{m}$ and compute original loss $\mathcal{L}_{\text{orig}}$
    \STATE Compute loss gap: $\Delta \mathcal{L} := \mathcal{L}_{\text{mask}} - \mathcal{L}_{\text{orig}}$
    \IF{$\Delta \mathcal{L} \leq 0$}
        \STATE Decrement patience counter: $\gamma \leftarrow \gamma - 1$
        \STATE Remove $e_{m}$ from the expert pool
        \IF{$\gamma = 0$}
            \STATE Stop expansion  
        \ENDIF
    \ELSE
        \STATE Retain $e_{m}$
    \ENDIF
\ENDIF
\end{algorithmic}
\end{algorithm}

Adaptive expansion terminates under either of the following conditions:
\begin{itemize}
    \item The number of experts reaches the maximum allowable value: $K = K_{\max}$.
    \item The previously added expert fails to improve the loss, as evaluated by negative log-likelihood (NLL) comparison at the following expansion step. Expansion is terminated once no improvement is observed over $\gamma$ times. See {Algorithm~\ref{alg:nllstopping}} for full details.
\end{itemize}

\noindent Once expansion stops satisfying due to either criterion, training during expansion phase continues as standard training with active redundancy regularization.
Restricting the expansion phase to the initial 10\% of training ensures training stability and minimizes the computational overhead associated with expansion algorithms, allowing the remainder of training to proceed with a static expert configuration. The full algorithm of \textsc{MASS} is summarized in Appendix A.

\subsection{Standard Training}
Following the adaptive expansion phase, the expert pool has converged to an optimal size with enough semantic coverage for training data. For the remaining 90\% of steps, the model is trained upon a classical gradient-guided training scheme. 

\begin{figure*}[t]
    \centering
    \includegraphics[width=0.92\textwidth]{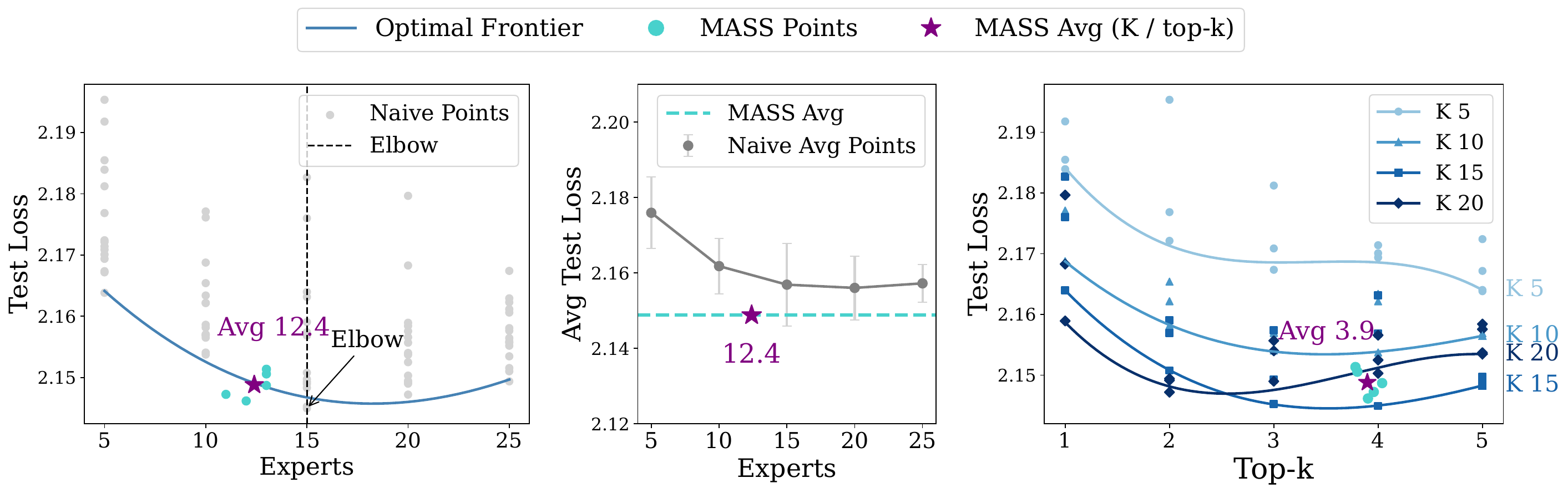}
    \caption{
    {Synthetic Results for Identifying Optimal MoE Configurations with \textsc{MASS}.}
    {Left:} Test loss vs.\ expert count for Naive and \textsc{MASS}. Gray dots represent individual runs of Naive with mixed top-$k$ across different \( K \), and the solid blue curve connects the best performing run per $K$, forming the optimal frontier. Elbow represented by a gray vertical dashed line denotes the point of the most cost-effective $K$. Green dots show five \textsc{MASS} runs, and the purple star shows their average (Avg 12.4 experts), which aligns closely with the elbow.
    {Center:} Average test loss comparison between Naive with varying expert counts (gray bars) and \textsc{MASS} average (dashed green line), showing consistently lower loss for \textsc{MASS} (Avg 2.15).
    {Right:} Test loss vs.\ top-\( k \) for fixed \( K \in \{5, 10, 15, 20\} \). Colored curves represent per-$K$ optimal frontier for Naive. The 5 \textsc{MASS} samples from the left plot are shared here as well (green circles), with the average active experts marked by the purple star (Avg 3.9).
    }
    \label{fig:syn_res3}
\end{figure*}

\begin{figure}[t]
    \centering
    \includegraphics[width=0.40\textwidth]{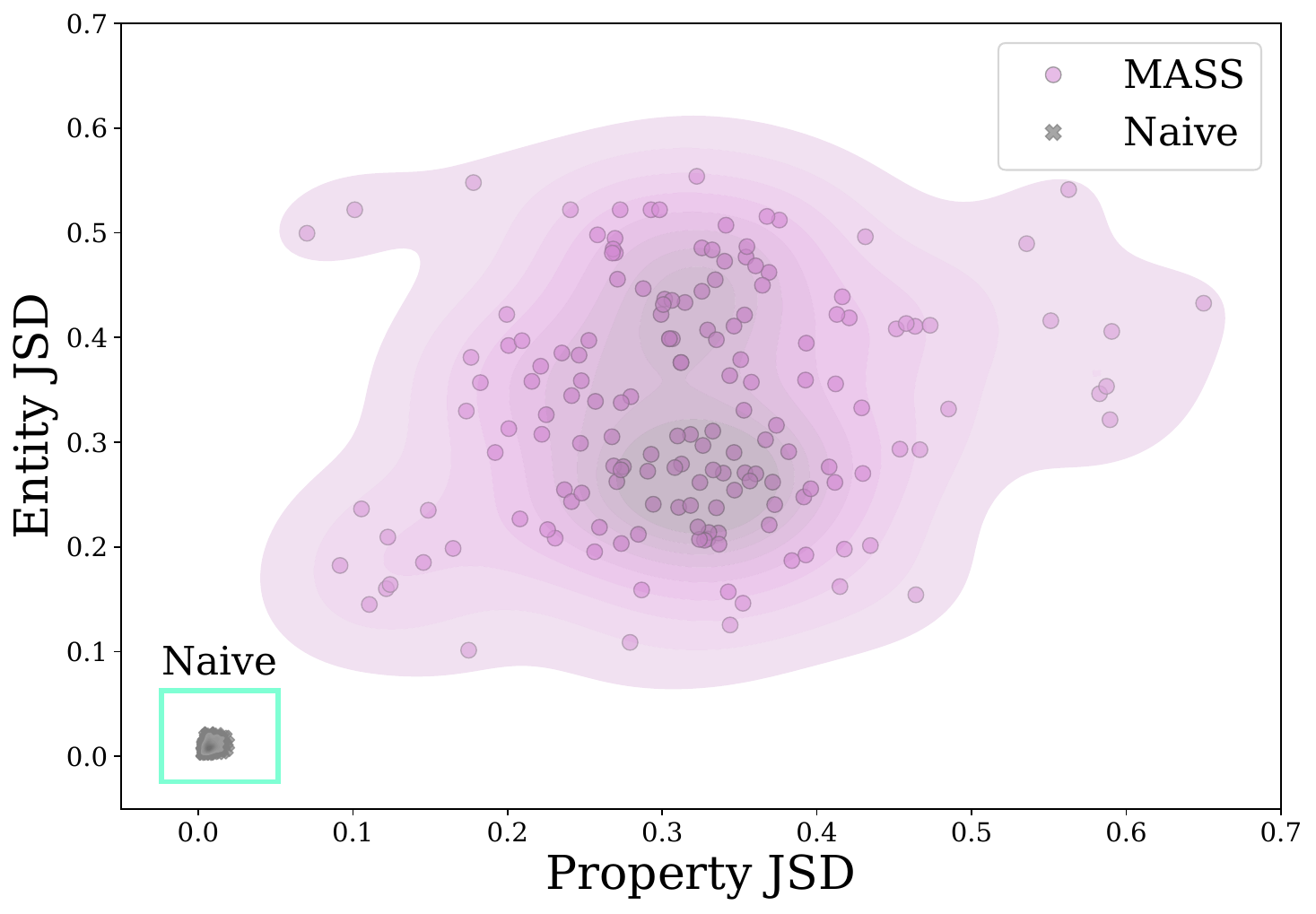}
    \caption{
    {Expert Specialization via Routing Divergence by Different Semantics.}
    Contour plot comparing pairwise JSD of average routing distributions grouped by semantic labels ($x$=entity and $y$=property) defined in the synthetic data. 
    }
    \vspace{-1em}
    \label{fig:syn_jsd}
\end{figure}

\section*{Results on Synthetic-Data}
We firstly evaluate \textsc{MASS} in a controlled synthetic setting based on a structured language modeling scenario. The use of a synthetic setup allows us to isolate the effects of adaptive expert expansion and dynamic routing, without confounding factors present in large-scale network and datasets. 

\subsection{Data Generation}
Here we describe full data generation process using multinomial Hidden Markov Models (HMMs), following the GINC dataset generation procedure~\cite{xie2022explanationincontextlearningimplicit}.
Each hidden state is a tuple $h_t = (v_t, s_t)$, where $v_t \in \mathcal{V}$ is an entity index and $s_t \in \mathcal{S}$ is a property type. A global memory matrix $M \in \mathcal{O}^{|\mathcal{V}| \times |\mathcal{S}|}$ deterministically maps each entity-property pair to a single token in the output vocabulary $\mathcal{O}$. The transition dynamics are defined as:
\begin{align*}
    s_{t+1} &\sim P(s_{t+1} \mid s_t; \theta), \quad \theta \sim p(\theta), \\
    v_{t+1} &\sim 0.9 \cdot \delta(v_{t+1} = v_t) + 0.1 \cdot I(v_{t+1} \mid v_t),
\end{align*}
where the model parameter $\theta \in \Theta$ serves as a concept parameter governing distinct property transitions, each sampled as a distinct HMM from a uniform mixture of HMMs over a family $\Theta$ of 5 concepts. The $\delta(v_{t+1} = v_t)$ denotes the Dirac delta function, assigning probability mass 1 to $v_{t+1} = v_t$ (i.e., entity persistence). This defines a sticky Markov process that maintains the same entity with high probability (0.9), and transitions to a new entity with a small probability (0.1) according to uniform $I(v_{t+1} \mid v_t)$. 
At each time step, the emitted token is deterministic as $o_t = M[v_t, s_t]$ with $p(o_t|(v_t, s_t)) = 1$. Let $T$ denote the total number of tokens in the generated sequence, $n$ the number of training examples, and $t'$ the length of each input sub-sequence $x_i$. Each example $(x_i, y_i)$ consists of an input $x_i = [o_{t_i}, o_{t_i + 1}, \dots, o_{t_i + t' - 1}]$ and a target token $y_i = o_{t_i + t'}$. Therefore, each training example consumes $t'+1$ tokens and total sequence length $T = n \cdot (t' + 1) + t'$. The final training sequence is represented as $[x_1, y_1, o_{\text{delim}}, \dots, x_n, y_n, o_{\text{delim}}, x_{\text{test}}]$
where $x_{\text{test}}$ is used to evaluate the model's prediction of $y_{\text{test}} = o_T$.

\subsection{Model}
For the synthetic experiment, we consider a single-layer Transformer architecture augmented with a Mixture-of-Experts (MoE) module. The model includes a token embedding layer, sinusoidal positional encoding, and a Transformer encoder block. The MoE module replaces the standard feed-forward network (FFN), where each expert is implemented as a single linear transformation followed by a \texttt{SiLU} activation, mapping from $\mathbb{R}^d$ to $\mathbb{R}^{d'}$. A gating network, implemented as a simple linear layer followed by a softmax activation, computes the routing score distribution over experts. The MoE output is computed as a weighted combination of routed expert outputs and passed through a shared decoder to produce token predictions. Models are trained end-to-end using cross-entropy loss.

\noindent We compare two models in this controlled setting:
\begin{itemize}
    \item \textbf{Naive MoE.} This baseline adopts a fixed number of experts, along with standard top-$k$ routing with fixed $k$.
    \item \textbf{\textsc{MASS}.} Sharing the same network architecture as Naive, it grows experts adaptively by expansion algorithm of \textsc{MASS} and uses {Top-$p$} gating.
\end{itemize}

\noindent This setup allows us to assess: 
(1) whether the \textsc{MASS} expansion algorithm successfully identifies an optimal expert pool size that balances semantic capacity and computational cost, and 
(2) whether \textsc{MASS}—via improved semantic differentiation and dynamic expert allocation —yields better performance compared to baseline.

\paragraph{Results}

We evaluate \textsc{MASS} against static MoE baselines by varying the number of experts $K \in \{5, 10, 15, 20, 25\}$ and top-$k \in \{1, 2, 3, 4, 5\}$ in Naive configurations, each repeated by three times. For each $K$, we select the best-performing Naive run with lowest test loss and connect these points to form the optimal frontier, a curve that represents the minimal test loss achievable per model capacity.
While this is not a strict Pareto front in the multi-objective optimization sense, it serves as an empirical optimality boundary for assessing the performance–cost trade-off. To ensure a fair comparison, we evaluate \textsc{MASS} on five independent runs.

\begin{figure*}[t]
    \centering
    \includegraphics[width=0.85\textwidth]{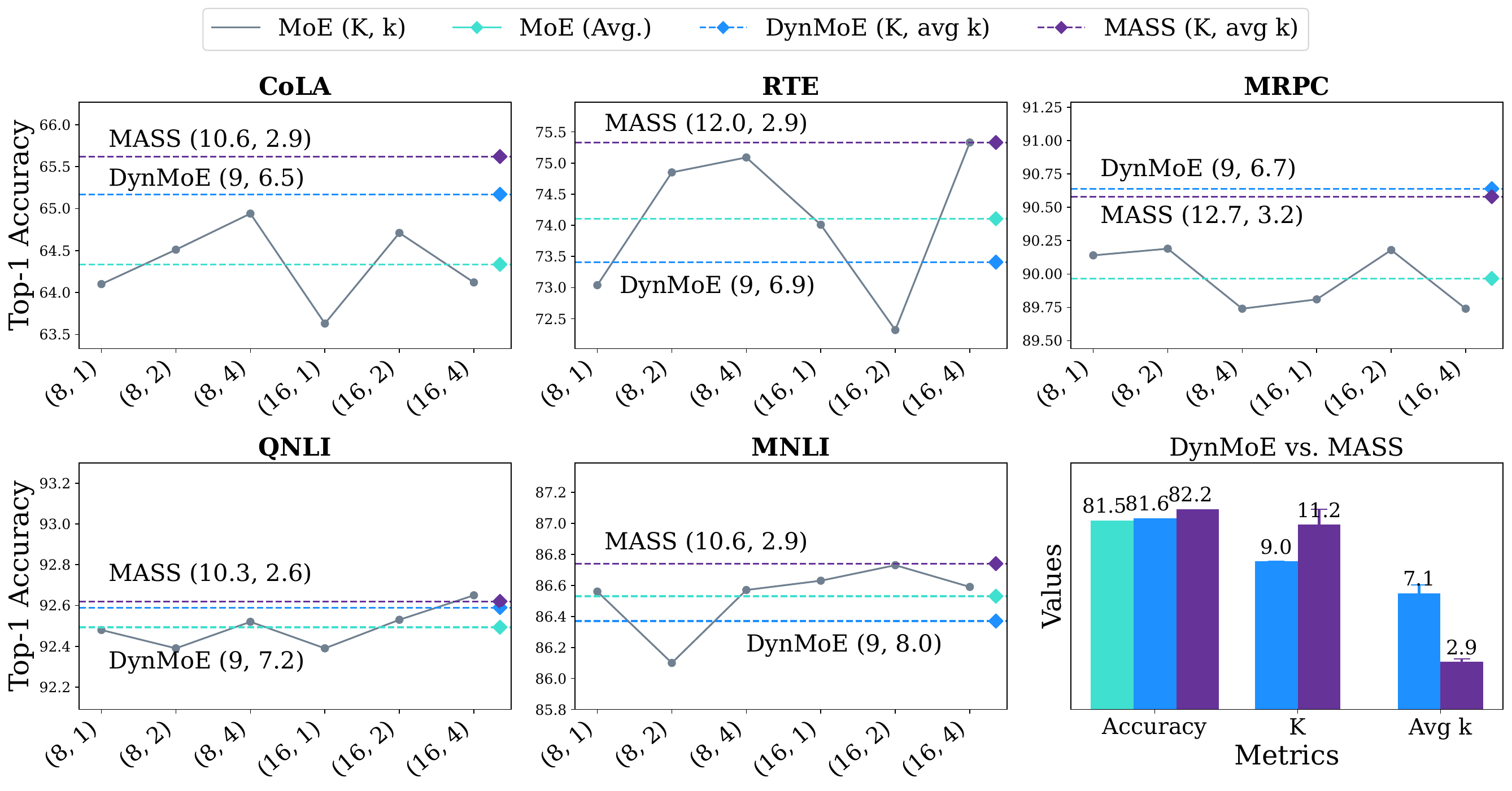}
    \caption{
    {Performance Comparison on GLUE benchmarks.}
    The top five plots show top-1 accuracy for each GLUE task comparing \textsc{MASS}, DynMoE, and static MoE configurations. Gray dots connected by gray line represent the performance of fixed MoE baselines, with horizontal lines marking average performance of DynMoE (blue) and \textsc{MASS} (purple). Numbers indicate the average expert count $K$ and activated $k$ across three random seeds.
    The bottom-right figure provides summary comparison for top-1 accuracy, $K$, and $k$ in average across models.
    }
    \label{fig:real_glue_res}
\end{figure*}

{Figure~\ref{fig:syn_res3}} summarizes the performance of \textsc{MASS} against these static MoE baselines. 
As shown in the empirical results in {left} plot that shows test loss as a function of expert counts, \textsc{MASS} effectively discovers cost-efficient expert configurations. Operating under dynamic expert expansion and routing, it consistently matches or surpasses the best-performing static MoE baselines. Importantly, \textsc{MASS} converges to expert counts near or below the empirical elbow ($K = 15$) with an average of 12.4 experts, demonstrating its ability to identify an optimal balance on cost-performance trade-off.
The {right} plot shows test loss versus top-$k$ for fixed $K \in \{5, 10, 15, 20\}$. \textsc{MASS} with {Top-$p$} routing achieves comparable or better results, with 3.9 active experts per token in average.
These results highlight that \textsc{MASS} not only discovers an effective expert pool size with enhanced semantic specialization, but also dynamically adapts expert allocation to equally distribute the routing confidence across all tokens.

We further analyze expert specialization behavior using the same set of sample runs for both Naive and \textsc{MASS}. For this, we analyze the diversity of expert routing conditioned on two semantic groups, entity $\in \mathcal{V}$ and property $\in \mathcal{S}$ of tokens in the synthetic data. Specifically, we compute the pairwise Jensen–Shannon Divergence (JSD) between the average expert routing distributions per entity (property) on each of the fully trained models, Naive and MASS. 
Formally, for a entity pair $(v_i, v_j)_{i,j \in \mathcal{V}}$, let $P_{v_i}, P_{v_j} \in \mathbb{R}^K$ denote the average routing distributions over $K$ experts for $v_i$ and $v_j$, respectively. The JSD is then computed as:
\[
\mathrm{JSD}(P_{v_i} \| P_{v_j}) = \frac{1}{2} \mathrm{KL}(P_{v_i} \| M) + \frac{1}{2} \mathrm{KL}(P_{v_j} \| M),
\]
where $M = \frac{1}{2}(P_{v_i} + P_{v_j})$ and $\mathrm{KL}$ denotes the Kullback–Leibler divergence.
This provides a symmetric and bounded measure in $[0, 1]$ of expert-entity (property) specialization. A larger JSD value implies that the routing distributions for varying input tokens are more diverging—the model assigns different subsets of experts to different semantics. In contrast, low JSD indicates that tokens having different semantics are routed similarly, a lack of specialization.

\begin{table*}[t]
\centering
\setlength{\arrayrulewidth}{0.0pt}
\begin{tabular}{@{\extracolsep{\fill}}l@{\hspace{1.0em}}!{\color{lightgray}\vrule}*{5}{c}}
\toprule
Algorithms & PACS & VLCS & OfficeHome & TerraIncognita & Average \\
\midrule
\rowcolor{white}
\arrayrulecolor{lightgray}
GMoE (K = 4) $^3$ & 88.2 & 79.8 & 73.5 & 47.8 & {72.3}\\
\rowcolor{white}
\arrayrulecolor{lightgray}
GMoE (K = 6) $^2$ & 88.1 & 80.2 & 74.2 & 48.5 & {72.8} \\
\rowcolor{white}
\arrayrulecolor{lightgray}
GMoE (K = 8) $^3$ & 88.2 & 80.0 & 74.2 & 47.2 & {72.4}
\\
\arrayrulecolor{lightgray}\midrule
\rowcolor{white}
\textit{Adaptive Experts} & & & & & \\
DynMoE (with Gshard Loss)$^1$ & 88.4 & 79.4 & 73.6 & 46.6$^3$ & {72.0} \\
\textbf{\textsc{MASS}} & 88.7 & 81.1 & 73.8 & 47.5 & {72.8} \\
\bottomrule
\end{tabular}
\caption{Domain Generalization Performance Comparison Across Multiple Visual Domains. Out-of-domain accuracies with train-validation selection criterion on four DomainBed datasets comparing fixed-size GMoE, DynMoE, and \textsc{MASS}. $^1$ from \citet{guo2025dynamicmixtureexpertsautotuning}, $^2$ from \citet{li2023sparsemixtureofexpertsdomaingeneralizable}, $^3$ reproduced by us (DynMoE result for TerraIncognita is not officially reported).}
\vspace{-1em}
\label{tbl:domainbed} 
\end{table*}

{Figure~\ref{fig:syn_jsd}} presents a contour plot of entity-level JSD (x-axis) vs.\ property-level JSD (y-axis), comparing \textsc{MASS} (purple) and Naive MoE (gray). 
As shown, \textsc{MASS} exhibits significantly higher JSD in both dimensions, forming a wide spread contour, whereas Naive MoE shows minimal variation clustered near zero in both axes. This suggests that \textsc{MASS} enables more distinct expert usage across different semantic groups, encouraging functional differentiation among experts. In contrast, Naive MoE uses experts uniformly, failing to separate semantic roles across experts. 

To verify the sanity of gradient shift detecting CPD algorithm used in \textsc{MASS}, we further provide a qualitative view of \textsc{MASS}’s internal dynamics by visualizing the gradient norm traces $\|\nabla_{\theta_k} \mathcal{L}_t\|_2$ for experts near their expansion events in {Appendix C}.

\section*{Results on Real-Data}
To assess the effectiveness of \textsc{MASS} in realistic settings, we evaluate its performance across two domains, natural language understanding and vision domain generalization. In both domains, we compare \textsc{MASS} against established MoE baselines, including fixed-size sparse MoEs, GMoE, and recent dynamic variant DynMoE.

\paragraph{Language Task}
For language task, we follow the MoEfication setup~\cite{zhang2022moeficationtransformerfeedforwardlayers} and fine-tune BERT-large~\cite{devlin2019bert} models on the GLUE benchmark~\cite{wang2019gluemultitaskbenchmarkanalysis}. Specifically, we evaluate on CoLA~\cite{warstadt2019neuralnetworkacceptabilityjudgments}, QNLI~\cite{wang2019gluemultitaskbenchmarkanalysis}, RTE~\cite{bentivogli2009fifth}, MNLI~\cite{xu-etal-2020-clue}, and MRPC~\cite{dolan-brockett-2005-automatically}, five subtasks of GLUE. We compare the performance of \textsc{MASS} with other baselines, MoE with cosine routers~\cite{li2023sparsemixtureofexpertsdomaingeneralizable} and DynMoE~\cite{guo2025dynamicmixtureexpertsautotuning}, with all results averaged over three random seeds. 

As shown in {Figure~\ref{fig:real_glue_res}}, \textsc{MASS} consistently matches or outperforms other MoE variants across all five GLUE tasks, achieving the highest average top-1 accuracy among the compared methods. Importantly, \textsc{MASS} adaptively expands its expert pool to a diverse range of sizes, $K \in [9.5, 12.7]$, depending on task complexity, while DynMoE consistently converges to a fixed size of $K=9.0$, showing limited responsiveness to task-specific demands.

Moreover, \textsc{MASS} exhibits a substantially lower average activated $k$ usage (2.6–3.2), activating only around 25\%–30\% of its expert pool per token. In contrast, DynMoE utilizes 70\%-90\% of its expert capacity with 6.5–8.0 active experts out of 9, reflecting denser and less specialized routing. This clear contrast in expert utilization pattern highlights \textsc{MASS}'s capacity to promote sparse, semantically differentiated expert activation, leading to more efficient and targeted expert usage across input tokens.

\paragraph{Vision Task}\label{subsec:rexp-vision}
For vision experiments, we follow the same MoE integration strategy in GMoE~\citep{li2023sparsemixtureofexpertsdomaingeneralizable}, applying it to a pre-trained ViT-S/16 model~\citep{dosovitskiy2021imageworth16x16words}. We evaluate performance on the DomainBed benchmark~\citep{gulrajani2020searchlostdomaingeneralization}, a widely used framework for domain generalization. DomainBed is constructed to assess a model's ability to generalize across multiple visual domains by training on a subset of domains and testing on an unseen target domain. We conduct experiments on four diverse datasets—PACS~\cite{li2017deeperbroaderartierdomain}, VLCS~\cite{albuquerque2021generalizingunseendomainsdistribution}, OfficeHome~\cite{venkateswara2017deephashingnetworkunsupervised}, and TerraIncognita~\cite{beery2018recognitionterraincognita}—each containing multiple domains with different environments.
In our setting, we adopt a leave-one-domain-out protocol: for each dataset, models are trained on all but one domain and evaluated on the held-out target. Model selection follows the standard train-validation scheme, where the checkpoint with the highest validation accuracy is used for evaluation. 

We compare \textsc{MASS} to GMoE with fixed expert sizes ($K = 4, 6, 8$) and DynMoE. Both \textsc{MASS} and DynMoE starts with the initial expert size 6. 
{Table~\ref{tbl:domainbed}} reports the domain generalization accuracy across four benchmarks in DomainBed. \textsc{MASS} consistently achieves competitive or superior performance compared to GMoE baselines. Compared to DynMoE, \textsc{MASS} delivers higher generalization across all domains. 
These results indicate that \textsc{MASS}'s adaptive expansion mechanism is effective beyond the language domain, offering improved generalization in vision tasks. 

\section*{Conclusion}

We introduced \textsc{MASS}, a framework for adaptive expert expansion in MoE models that promotes semantic specialization through gradient-guided expansion and dynamic routing.
\textsc{MASS} delivers three key benefits: (1) it discovers an optimal expert pool size without requiring costly hyperparameter search over $K$, (2) it enhances MoE performance by promoting semantic differentiation across experts, reducing functional redundancy and enabling more complementary expert behaviors, achieving superior performance even with fewer active experts per token, and (3) it ensures balanced expert attention per token via adaptive expert selection based on routing confidence mass. These advantages are demonstrated across synthetic, language, and vision benchmarks, confirming both the robustness and generality of the approach. 

\section*{Acknowledgements}

This work was partly supported by the Institute for Information \& Communications Technology Planning \& Evaluation (IITP) grants funded by the Korean government (MSIT) (No. RS-2022-II220113, Developing a Sustainable Collaborative Multi-modal Lifelong Learning Framework; No. RS-2025-25442149, LG AI STAR Talent Development Program for Leading Large-Scale Generative AI Models in the Physical AI Domain), and Samsung Research Funding \& Incubation Center of Samsung Electronics under Project Number SRFC-IT2402-08.

\bibliography{aaai2026}

\end{document}